\documentclass[journal,twoside,web]{ieeecolor}
\pdfoutput=1
\usepackage{etoolbox}
\makeatletter
\@ifundefined{color@begingroup}%
{\let\color@begingroup\relax
\let\color@endgroup\relax}{}%
\def\fix@ieeecolor@hbox#1{%
\hbox{\color@begingroup#1\color@endgroup}}
\patchcmd\@makecaption{\hbox}{\fix@ieeecolor@hbox}{}{\FAILED}
\patchcmd\@makecaption{\hbox}{\fix@ieeecolor@hbox}{}{\FAILED}

\usepackage{tmi}
\usepackage{cite}
\usepackage{amsmath,amssymb,amsfonts}
\usepackage{algorithmic}
\usepackage{graphicx}
\usepackage{comment}
\usepackage{textcomp}
\usepackage{subfigure}
\usepackage{multirow}
\usepackage{makecell}
\usepackage{booktabs}
\usepackage{color}
\usepackage{verbatim}

\usepackage{hyperref}

\hypersetup{colorlinks,
     citecolor=green,
     filecolor=blue,
     linkcolor=blue,
     urlcolor=cyan
}

\usepackage[T1]{fontenc}
\usepackage{listings}

\def\BibTeX{{\rm B\kern-.05em{\sc i\kern-.025em b}\kern-.08em
    T\kern-.1667em\lower.7ex\hbox{E}\kern-.125emX}}
\begin{document}
\title{Learning with Explicit Shape Priors for Medical Image Segmentation}
\author{Xin You, Junjun He, Jie Yang \IEEEmembership{Member, IEEE}, Yun Gu \IEEEmembership{Member, IEEE}
\thanks{X. You, J. Yang and Y. Gu are with the Institute of Image
Processing and Pattern Recognition, Shanghai Jiao Tong University,
Shanghai 200240, China. }
\thanks{J. He is with Shanghai AI Laboratory, Shanghai 200240, China.}
}

\maketitle

\begin{abstract}
Medical image segmentation is a fundamental task for medical image analysis and surgical planning. In recent years, UNet-based networks have prevailed in the field of medical image segmentation. However, convolution-neural networks (CNNs) suffer from limited receptive fields, which fail to model the long-range dependency of organs or tumors. Besides, these models are heavily dependent on the training of the final segmentation head. And existing methods can not well address these two limitations at the same time. Hence, in our work, we proposed a novel shape prior module (SPM), which can explicitly introduce shape priors to promote the segmentation performance of UNet-based models. The explicit shape priors consist of global and local shape priors. The former with coarse shape representations provides networks with capabilities to model global contexts. The latter with finer shape information serves as additional guidance to boost the segmentation performance, which relieves the heavy dependence on the learnable prototype in the segmentation head. To evaluate the effectiveness of SPM, we conduct experiments on three challenging public datasets. And our proposed model achieves state-of-the-art performance. Furthermore, SPM shows an outstanding generalization ability on classic CNNs and recent Transformer-based backbones, which can serve as a plug-and-play structure for the segmentation task of different datasets. Source codes are available at \url{https://github.com/AlexYouXin/Explicit-Shape-Priors}

\end{abstract}

\begin{comment}
Medical image segmentation is a fundamental task for medical image analysis and surgical planning. Previous works attempted to incorporate shape priors into segmentation models, which is beneficial to attain finer prediction of anatomical structures. In this work, we detailedly discuss three types of segmentation paradigms with shape priors, including atlas-based models, statistical shape models and UNet-based models. The former two types of models show a poor generalization ability, thus UNet-based models have dominated the field of medical image segmentation in recent years. However, existing UNet-based models mainly employ implicit shape priors, which do not have a good interpretability and generalization ability on different organs with distinctive shapes. Thus, we proposed a novel shape prior module (SPM), which can explicitly introduce shape priors to promote segmentation performance of UNet-based models. To evaluate the effectiveness of SPM, we conduct experiments on three challenging public datasets. And our proposed model achieves state-of-the-art performance. Furthermore, SPM shows an outstanding generalization ability on classic convolution-neural-networks (CNNs) and recent Transformer-based backbones, which can serve as a plug-and-play structure for the segmentation task of different datasets. Codes are available at \url{https://github.com/AlexYouXin/Explicit-Shape-Priors}
\end{comment}


\section{Introduction}
\label{sec:introduction}
Medical image segmentation is regarded as one of the most essential and challenging tasks for medical image analysis, which is a prerequisite for  image-guided diagnosis and computer-assisted intervention \cite{hesamian2019deep}.  It provides anatomical shape information by making per-pixel predictions for organs or lesions in images \cite{azad2022medical}. Recently, deep learning techniques \cite{shen2017deep} have dominated medical image segmentation. The most successful architectures are U-shape networks \cite{ronneberger2015u}. The multi-scale features are extracted for specific regions in the encoder, which contain semantic and detailed information. Then deep features from the bottleneck are fused with encoded features via skip connections in the decoder structure. Lastly, per-pixel classifications are carried out on decoded features via the segmentation head \cite{zhou2022rethinking}, which is a learnable prototype. Deep networks are free from hand-crafted features to achieve outstanding performance for segmentation. 

\begin{figure*}[!t]
\centerline{\includegraphics[width=0.9\linewidth]{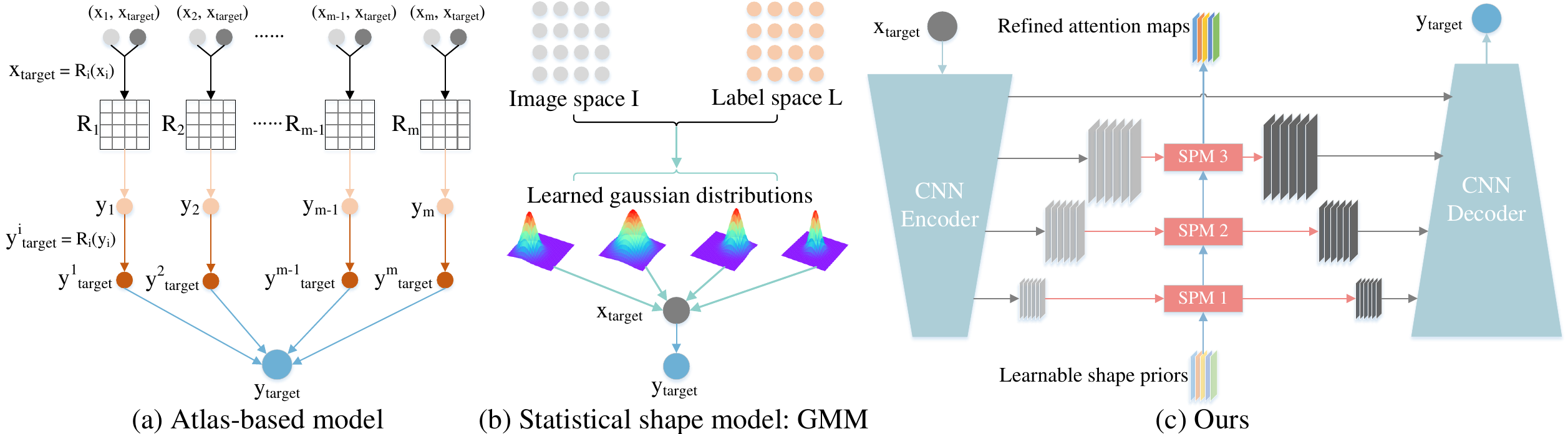}}
\caption{The comparison between different segmentation paradigms with explicit shape priors. (a) Atlas-based models employ m ground truths from target datasets as shape priors, which construct the transformation matrix $\mathcal{R}$ between source and target images, then apply $\mathcal{R}$ to shape priors to achieve segmentation masks. (b) Gaussian Mixture Model (GMM) gathers N (the number of segmentation classes) learnable Gaussian distributions as shape priors to deal with the per-pixel classification. (c) Our model adopts N-channel learnable shape priors as additional inputs to boost the segmentation performance of deep neural networks.}
\label{segmentation paradigm}
\end{figure*}

\begin{comment}
    Then decoded features with a size of $H \times W \times D$ are mapped into $H \times W \times N$, which needs $D \times N$ parameters for prototype learning. However, this operation hurts generalization abilities.

    Besides, decoded features with a size of $H \times W \times D$ are mapped into $H \times W \times N$, which needs $D \times N$ parameters for prototype learning. However, this operation hurts generalization abilities.
\end{comment}

\subsection{Limitations}
However, current UNet-based models suffer from the following limitations for medical image segmentation. \normalsize{\textcircled{\scriptsize{1}}}\normalsize CNNs bear limited receptive fields due to intrinsic properties of convolutional kernels, which cannot exploit long-range and global spatial relations between organs or tissues, then they fail to achieve fine shape representations. Thus, some implicit attention modules \cite{el2020bb, woo2018cbam, hu2018squeeze, oktay2018attention} are employed to enlarge models' receptive fields, and they are termed implicit shape models in our work. We will talk about them comprehensively in Section \ref{Implicit shape model}. \normalsize{\textcircled{\scriptsize{2}}}\normalsize Segmentation masks are primarily based on the training of the final learnable prototype \cite{zhou2022rethinking} interpreted by the segmentation head. Specifically, considering a segmentation task with $N$ semantic classes, $N$ class-wise prototypes are learned for pixel-wise classification. Only one learnable prototype is learned for each class, which employs limited representation abilities, thus insufficient to describe rich intra-class variance. Under this circumstance, UNet-based models meet a major challenge to extract precise shape information of organs or tumors. Specific loss functions \cite{el2021high, ma2021loss} are designed to integrate explicit shape priors or anatomical constraints to segmentation frameworks instead of Dice loss or cross-entropy loss, which can extract sufficient structure information related to the regions of interest, including shapes and topology. However, these loss functions are task-specific and cannot be easily extended on different datasets. Moreover, explicit shape models \cite{kalinic2009atlas, heimann2009statistical} are proposed to enhance models' capacities for shape representations, with shape priors as an additional input. More thorough descriptions can be referred to in Section \ref{Explicit shape model}.

\subsection{Implicit Shape Models}\label{Implicit shape model}
To address the limitation on restricted receptive fields of CNNs, previous works try to introduce implicit anatomical shape priors to the U-shape structures, which are termed implicit shape models. These shape priors can be injected into network structures via implicit attention modules. Theoretically, attention modules $\mathcal{M}$ \cite{el2020bb, woo2018cbam, hu2018squeeze, oktay2018attention, chen2021transunet} are introduced to strengthen deep features generated from encoders $\mathcal{E}$, driving them more focused on foreground regions with specific shapes. Then decoders $\mathcal{D}$ merge enhanced features to attain more precise masks $\mathcal{Y}$ of corresponding images $\mathcal{X}$. This process can be described as follows:
\begin{eqnarray}
  & \mathcal{Y} = \mathcal{D}(\mathcal{M}(\mathcal{E}(\mathcal{X})))
  \label{implicit shape models}
\end{eqnarray}
More specifically, the attention modules can be divided into two categories. The first type belongs to convolution-based attention modules. BB-UNet \cite{el2020bb} enhanced skipped features via bounding box (BB) filters generated before training. Though BB filters can provide shape information of specific organs, obtaining BB filters requires manual interventions. Besides, Attention UNet \cite{oktay2018attention} employs attention gates (AGs) to enhance salient features beneficial for specific tasks. AGs can also suppress redundant feature activation from irrelevant regions, which bear shape priors to some extent. CBAM \cite{woo2018cbam} adopted the channel attention module and spatial attention module to boost representation abilities on the shape of specific regions. However, stacking convolution-based attention modules cannot efficiently broaden effective receptive fields \cite{luo2016understanding}. And they still show limited competence to model long-term dependency. 

Different from that, the second type of attention module is based on a self-attention mechanism \cite{dosovitskiyimage}, which provides a feasible way of modeling global contexts via query, key, and value vectors. Many Transformer-based models with various types of self-attention \cite{chen2021transunet, tang2022self, valanarasu2022unext, hatamizadeh2022unetr, zhou2021nnformer} are proposed to model the long-range dependency of medical images. TransUNet \cite{chen2021transunet} combines 2D UNet with a pre-trained Vision Transformer (ViT) to solve volumetric image segmentation by stacking each slice's prediction. SwinUNETR \cite{tang2022self} adopts the shifted window-based attention to extract features of 3D patches, then merge multi-scale encoded features via residual convolutional blocks to attain final masks. However, unlike CNNs, these Transformer-based models require large data resources for training, which fail to simply and finely learn inductive bias such as shape prior information inside data sources \cite{xu2021vitae}.

\begin{comment}
    Different from the implicit shape priors mentioned above, some methods enhanced segmentation models with general explicit shape priors.
\end{comment}

\subsection{Explicit Shape Models}\label{Explicit shape model}
To relieve the heavy dependence on the training of the final learnable prototype, prior methods attempt to introduce additional shape information into the segmentation framework, which we call explicit shape priors. Different from implicit shape priors mentioned above, explicit shape priors show strong interpretability, which presents a rough localization for the regions of interest (ROIs). These works can be divided into two categories, including atlas-based models \cite{kalinic2009atlas}, and statistical shape models, represented by Gaussian Mixture Model (GMM) \cite{tai2005local}.

% and UNet-based models with specifically designed shape priors

\begin{comment}
  \begin{figure}[!t]
\centerline{\includegraphics[width=\columnwidth]{introduction for segmentation mask.pdf}}
\caption{Visualizations on images and ground truths of the BraTS 2020, ACDC, and VerSe 2019 datasets.}
\label{segmentation mask}
\end{figure}  
\end{comment}

\begin{comment}
% , which is also the core step of image registration \cite{cabezas2011review}
Atlas-based segmentation models indeed extract intensity-based transformation relations $R_{I}$ between source images in atlas and target images. Then after applying $R_{I}$ transform to source ground truths (GTs), we can attain GTs of target images. However, the registration process can be very time-consuming, and the choice of source images used for registration can be very tricky \cite{kalinic2009atlas}.
\end{comment}

% , from training and testing datasets respectively
Specifically, the first paradigm is based on the atlas, whose essence is the label propagation via registration transform between source and target images \cite{cabezas2011review}. Then after applying this transform to source ground truths (GTs), we can attain GTs of target images. Obviously, nonrigid registration cannot perfectly deal with the segmentation task due to limited data sources and imaging noise \cite{bach2015atlas}. Thus, a more feasible solution is to build matching relations in a non-local way. Besides, it is beneficial to achieve a more refined segmentation mask by adopting the weighted combination of a group of candidates from source images, which are called registration bases. The segmentation masks of registration bases serve as shape priors to promote the segmentation of target images. The whole model can be described by Eq. \ref{atlas-based models}:

\begin{scriptsize}
\begin{eqnarray}
  & \mathcal{Y}_{test} = \sum \limits_{i=1}^{m} \omega_{i} \times \mathcal{T}(\mathcal{Y}_{b}^{i}; \mathcal{R}(\mathcal{X}_{b}^{i} \Rightarrow \mathcal{X}_{test})), \mathcal{X}_{b}^{i} \in \mathcal{X}_{train}, \mathcal{Y}_{b}^{i} \in \mathcal{Y}_{train}
  \label{atlas-based models}
\end{eqnarray}
\end{scriptsize}

% , $m$ is the dimension of registration space
where $\mathcal{X}_{train}$ and $\mathcal{Y}_{train}$, $\mathcal{X}_{test}$ and $\mathcal{Y}_{test}$ represent images and GTs from the training and testing data sources, $\mathcal{X}_{b}$ and $\mathcal{Y}_{b}$ represent the group of registration bases containing $m$ pairs of source images and GTs, $\omega_{i}$ is the weighting coefficient. Besides, $\mathcal{R}$ refers to the registration transform between $\mathcal{X}_{b}^{i}$ and $\mathcal{X}_{test}$. And $\mathcal{T}$ is the transformation matrix, which applies the registration transform to each GT of registration bases $\mathcal{Y}_{b}^{i}$.

% If one case shows little similarity to registration bases, then we cannot gain a fine predicted mask. 
% Furthermore, there exists a large computational cost for the inference process due to the high complexity of atlas-based models.

For atlas-based models, there exists a large computational cost during inference. Furthermore, the choice of registration bases is significant to the model's robustness. Detailedly, the base vector should cover the distribution property of the whole dataset. However, adopting fixed template shapes cannot cover all biological objects due to the existence of shape variations. Thus, gathering a number of shape priors in a statistical way from training datasets is essential to boost models' segmentation performance and robustness. 

\begin{comment}

The second category is statistical shape models \cite{heimann2009statistical}. Of all statistical-based models, GMM \cite{reynolds2009Gaussian} is a typical model which extracts a batch of Gaussian distributions with learnable means and variances. And these distributions can serve as explicit shape priors representing shape characteristics of training datasets. Nevertheless, these models can provide a inconsistent segmentation mask as a result of imaging noise. Fixed Gaussian distributions are not robust enough to cover a wide range of datasets. Thus, there may be poor segmentation results for unseen datasets. %  \cite{dar2019medical}
    
\end{comment}

The second segmentation paradigm is statistical shape models \cite{heimann2009statistical}. A representative method is the Gaussian Mixture Model (GMM) \cite{reynolds2009Gaussian}, which completes a consecutive mapping from image space $I$ to label space $L$ via a group of learnable Gaussian distributions. And these Gaussian distributions can be regarded as explicit shape priors of the dataset. During training, Expectation Maximization (EM) algorithm is iteratively implemented to update the learnable Gaussian distributions and segmentation masks \cite{tai2005local}. In the inference process, we utilize learned shape priors as independent kernels. The whole model is illustrated by the following equation:
\begin{eqnarray}
  & \mathcal{Y}_{test} = \mathop{\arg\max} \limits_{i=1,...,N} \mathcal{G} (\mathcal{X}_{test}; \mathcal{K}_{i}), \mathcal{K}_{i} \in \mathcal{K}(\mathcal{X}_{train}, \mathcal{Y}_{train})
  \label{GMM models}
\end{eqnarray}

Where $\mathcal{K}$ means learned Gaussian distributions generated from the training process, $\mathcal{K}_{i}$ refers to each element in $\mathcal{K}$, $N$ is the number of Gaussian kernels, which is also the number of semantic classes. $\mathcal{G}$ is a mapping function by distributing n Gaussian probability values to each pixel, each value generated from kernel $\mathcal{K}_{i}$. However, GMM is still sensitive to noises and dynamic backgrounds. Besides, the initial setting of the EM algorithm is crucial to the final solution \cite{reynolds2009Gaussian}.
% , and a poor initial value tends to achieve a local optimal solution 

% Cootes et al. \cite{cootes1995active} proposes the active shape model (ASM), which trains a prior model, then attempts to search the most similar match between the model and new data.
Some other related works expand the impacts of statistical shape models. Kass et al. \cite{kass1988snakes} implement the contour of images with an explicit curve by locally optimizing the external and internal energy functions. But this explicit representation has trouble in dealing with multiple objects and the final results are closely bound to the initialization. To solve these limitations, some work \cite{schnorr1992computation, aubert2003image} tries to fit
statistical models according to intensity, color, texture, or motion inside separated regions, instead of local gradient information. However, these explicit shape models cannot be embedded into deep segmentation networks, which still suffer from poor generalization ability for unseen datasets.

% Experiments demonstrate that SPM can boost the segmentation performance of various models on three datasets above.
\subsection{Contributions}
To address the mentioned two limitations in the meantime, we incorporate learnable explicit shape priors to enhance shape representations of UNet-based models. In the field of object detection, DETR \cite{carion2020end} introduced a set of learnable object queries, then reasons about the relations of the objects and the global image context to directly output the final predictions. Motivated by this design, we devise the learnable shape prior, which is indeed an N-channel vector, and each channel contains rich shape information for the specific class of regions. Shape priors are generated based on self-attention, which endow the segmentation model with global receptive fields. Meanwhile, learnable shape priors can boost encoded deep features with richer shape information, then drive networks to generate better masks, which can ease the heavy dependence on the learnable prototype. Also, encoded features will contribute to iterative updates of shape priors. Based on this theory, we propose the shape prior module (SPM) consisting of the self-update block and cross-update block, placing it in the structure of skip connections as shown in Figure \ref{segmentation paradigm}. SPM is evaluated on BraTS 2020, ACDC, and VerSe 2019. And our model achieves state-of-the-art (SOTA) segmentation performance on these datasets. Besides, due to its plug-and-play property, we probe into the generalization ability on other networks, including classic CNNs and recent Transformer-based models. Our contributions are summarized as follows:

1) We conduct a thorough comparative analysis of three types of segmentation paradigms with explicit shape priors.

2) We propose a novel shape prior module (SPM), which consists of the self-update and cross-update block.

3) The proposed module helps us achieve SOTA performance on BraTS 2020, VerSe 2019, and ACDC.

4) SPM is a plug-and-play structure, which brings a significant boost to classic CNNs and Transformer backbones.

\section{Methodology}
\subsection{Unified Framework for Explicit Shape Models}\label{different paradigms about shape priors}
As shown in Figure \ref{segmentation paradigm}, we mainly discuss three types of segmentation paradigms, which can provide explicit shape priors. These paradigms can be unified as follows:
\begin{eqnarray}
  & \mathcal{O} = \mathcal{D}(\mathcal{P}(\mathcal{I}; \mathcal{S}))
  \label{uniform manner}
\end{eqnarray}
where $\mathcal{I}$ represents testing images as the input of the segmentation framework, and $\mathcal{O}$ refers to the model's outputs. $\mathcal{S}$ denotes explicit shape priors generated with different manners, which are used for enhancing the segmentation performance. $\mathcal{P}$ refers to the process of model prediction with joint inputs. $\mathcal{D}$ means the one-hot decoding on the generated N-channel prediction, and N is the number of segmentation classes.

In this work, the proposed paradigm introduces learnable explicit shape priors $\mathcal{S}$ to U-shape neural networks. Specifically, $\mathcal{S}$ is utilized as inputs of networks combined with images. The outputs of networks are predicted masks and attention maps generated by $S$. Then channels of attention maps can provide rich shape information of the ground truth region. The explicit-shape-prior model can be depicted as follows:
\begin{eqnarray}
  & \mathcal{Y}_{test}, attention = \mathcal{F}(\mathcal{X}_{test}, \mathcal{S}(\mathcal{X}_{train}, \mathcal{Y}_{train}))
  \label{shape prior}
\end{eqnarray}
where $\mathcal{F}$ represents the forward propagation during inference, $\mathcal{S}$ stands for consecutive shape priors constructing the mapping between image space $I$ and label space $L$. Here $\mathcal{S}$ is updated in the training process as the image-GT pair varies. Once training is finished, learnable shape priors are fixed, which can dynamically generate refined shape priors as input patches vary in the inference process. And refined shape priors serve as attention maps, which can localize regions of interest, and suppress background areas. Furthermore, a small portion of inaccurate ground truths will not affect the learning for $\mathcal{S}$ significantly, revealing the robustness of our proposed paradigm.

\subsection{Shape Prior Module}
\noindent\textbf{Overview.} As depicted in Figure \ref{segmentation paradigm}, our proposed model is a hierarchical U-shape network, which consists of a ResNet-like encoder, a Resblock-based \cite{tang2022self} decoder and the shape prior module (SPM). And SPM is a plug-and-play module, which can be flexibly plugged into other network structures to improve segmentation performance. In the sections below, we will give a detailed description of SPM, including the motivation for this module, the detailed structure, and the functions.

\begin{figure}[!t]
\centerline{\includegraphics[width=\columnwidth]{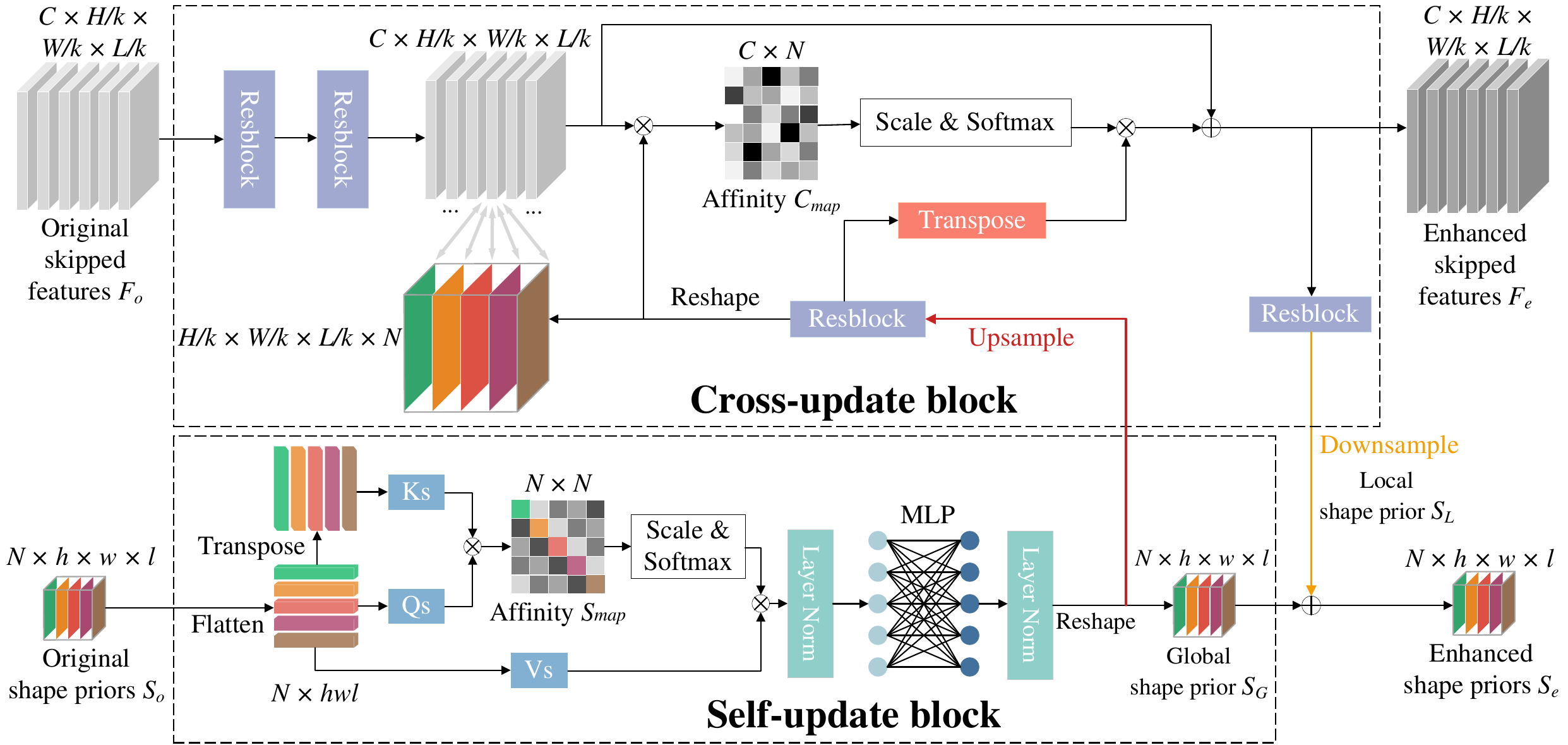}}
\caption{Illustration of the shape prior module (SPM). SPM consists of the self-update block (SUB) and cross-update block (CUB). Specifically, SUB aims to extract global shape priors via a linear attention module. CUB introduces global shape priors as additional guidance to refine original skipped features $F_{o}$. Also, the convolutional feature guides the generation of local shape priors with finer shape information. Here $H \times W \times L$ is the input patch size (k = 2, 4, 8), and $h \times w \times l$ is $\tfrac{1}{16}$ of the patch size.}
\label{SPM}
\end{figure}

% We can have a clear visualization on features before and after being processed by SPM in Figure \ref{feature enhance: brats and acdc}
In order to get rid of the dependency on the final learnable prototype, we propose to introduce explicit shape priors to UNet-based networks, which will exert anatomical shape constraints for each class to enhance the representation abilities of networks. Motivated by DETR \cite{carion2020end}, we devise n (the number of segmentation classes) learnable prototypes, the analogy to object queries in the Transformer decoder of DETR. As shown in Figure \ref{SPM}, inputs of SPM are original skipped features $F_{o}$ and original shape priors $S_{o}$, which are refined as enhanced skipped features $F_{e}$ and enhanced shape priors $S_{e}$. Specifically, learnable shape priors will generate refined attention maps with sufficient shape information under the guidance of convolutional encoded features. In the meantime, encoded features will generate more accurate segmentation masks via shape priors. Different from DETR, SPM will interact with multi-scale features, not just features from the bottleneck of the encoder. Thus, hierarchical encoded features before skip connections will be equipped with richer shape information via SPM. Enhanced shape priors are made up of two components, global and local shape priors, generated from the self-update block and cross-update block respectively. We will give a more elaborate description of these two blocks.

\noindent\textbf{Self-Update Block: Modeling long-range dependency.}
On the ground that we aim to introduce explicit shape priors which can localize the target regions, the size of shape priors $\mathcal{S}_{o}$ is $N$ $\times$ spatial dimension. $N$ refers to the number of classes, and spatial dimension is related to the patch size. To alleviate the drawback of limited receptive fields,  the long-range dependency inside shape priors is considered in this work. Specifically, the self-update block (SUB) is proposed to model relations between inter-classes and generates global shape priors with interactions between N channels. Motivated by the self-attention mechanism of Vision Transformer (ViT) \cite{dosovitskiyimage}, the affinity map of self-attention $S_{map}$ between $N$ classes is constructed by the Eq. \ref{affinity map}, which describes the similarity and dependency relationship between each channel of shape priors.
\begin{eqnarray}
  & S_{map} = Softmax(\frac{Q_{s}(\mathcal{S}_{o}) \times K_{s}(\mathcal{S}_{o})^{T} }{\sqrt{N}})
  \label{affinity map}
\end{eqnarray}
where $Q_{s}$ and $K_{s}$ represent convolutional transforms which project $\mathcal{S}_{o}$ into the query and key vector, $T$ is the transpose operator, and the dimension for $S_{map}$ is $N \times N$. The vanilla self-attention module shows quadratic computational complexity, which poses an obstacle to dense prediction tasks. Thus, many related works \cite{chen2021crossvit} attempt to reduce the computational cost of the self-attention module for faster convergence and less memory consumption. Here we set the spatial dimension of $\mathcal{S}_{o}$ as $h \times w \times l$, $\tfrac{1}{16}$ ratio of the patch size $H \times W \times L$. Besides, $\mathcal{S}_{o}$ bears $N$ tokens, which means the self-attention module in SUB is a linear attention module. And Eq. \ref{affinity map} shows $\mathcal{O}(N^{2} \times HWL)$ computational complexity.

After that, the weighted sum of $S_{map}$ and value vector of $\mathcal{S}_{o}$, noted as $V_{s}(\mathcal{S}_{o})$, are adopted to obtain global shape priors. This process also requires $\mathcal{O}(N^{2} \times HWL)$ computational costs. To further model the long-range dependency inside learnable shape priors, multi-layer perceptron (MLP) and layer normalization (LN) are employed. The detailed process can be illustrated as follows: 
\begin{eqnarray}
  & \mathcal{S'} = LN(S_{map} \times V_{s}(\mathcal{S}_{o})) + \mathcal{S}_{o} \\
  & \mathcal{S_{G}} = LN(MLP(\mathcal{S'})) + \mathcal{S'}
  \label{self attention}
\end{eqnarray}
where $V_{s}$ represents the convolutional transform, $\times$ means the process of matrix multiplication, $\mathcal{S_{G}}$ is global shape priors. Detailedly, $\mathcal{S_{G}}$ can build the long-term dependency inside $\mathcal{S}_{o}$, which contains global contexts of sampled input patches, including coarse shape and positional representations combined with sufficient texture information for global regions.
%  We will give a detailed visualization of $\mathcal{S_{G}}$ in the section \ref{visualization section}.

% \textcolor[RGB]{19,138,218}
\noindent\textbf{Cross-Update Block: Modeling local shape priors.} To relieve the dependence on the learnable prototype, we attempt to introduce explicit shape priors to boost the representation abilities for shape information. However, the structure of SUB falls lack of inductive bias \cite{xu2021vitae} to model local visual structures and localize objects with various scales. As a result of that, global shape priors do not have precise shape and contour information. Further, models have to learn intrinsic inductive bias from large amounts of data for a longer training time. To address this limitation, we propose the cross-update block (CUB). Motivated by the fact that convolutional kernels intrinsically bear the inductive bias of locality and scale invariance, CUB based on convolution injects inductive bias to SPM for local shape priors with finer shape information. Moreover, based on the fact that convolutional features from the encoder have remarkable potentials to localize discriminative regions \cite{zhou2016learning}, we attempt to interact original skipped features $F_{o}$ from the backbone with shape priors $\mathcal{S}_{o}$ as demonstrated in Figure \ref{SPM}. 

Specifically, we calculate the similarity map between features $F_{o}$ and shape priors $\mathcal{S}_{o}$. Here the dimension of $F_{o}$ is $C \times \tfrac{H}{k} \times \tfrac{W}{k} \times \tfrac{L}{k} (k = 2, 4, 8)$, and C represents the channel number of features. However, $F_{o}$ and $\mathcal{S}_{o}$ bear different scales, which makes it difficult to fuse two elements. Thus, we firstly upsample $\mathcal{S}_{o}$ to the same resolution as $F_{o}$, then integrate them based on the cross attention mechanism \cite{carion2020end}. The detailed computational process is illustrated as Eq. \ref{cross}:
\begin{eqnarray}
  & C_{map} = Softmax(\frac{Q_{c}(F_{o}) \times K_{c}(Upsample(\mathcal{S_{G}}))^{T} }{\sqrt{N}})
  \label{cross}
\end{eqnarray}
where $C_{map}$ means the affinity map in the cross attention stage, $Q_{c}$ and $K_{c}$ represent convolutional transforms which project $F_{o}$ and $\mathcal{S}_{o}$ into the query and key vector. $C_{map}$ is a 
$C \times N$ matrix, which evaluates the relations between $C$-channel feature maps $F_{o}$ and $N$-channel shape priors. The specific channel of convolutional feature maps $F_{o}$ correlates with specific channels of shape priors. After that, $C_{map}$ acts on transformed global shape priors $\mathcal{S_{G}}$ to refine $F_{o}$, with more accurate shape characteristics and rich global textures.
\begin{eqnarray}
  & F_{e} = C_{map} \times V_{c}(Upsample(\mathcal{S_{G}})) + F_{o}
  \label{enhanced skipped feature}
\end{eqnarray}
where $V_{c}$ refers to convolutional transforms projecting $\mathcal{S}_{o}$ into the value vector, $F_{e}$ represents enhanced skipped features. At the same time, local shape priors $\mathcal{S_{L}}$ are generated from downsampled $F_{e}$, which bear the property to model local visual structures (edges or corners).
\begin{eqnarray}
  & \mathcal{S_{L}} = Downsample(Conv(F_{e})) \\
  & \mathcal{S}_{e} = \mathcal{S_{L}} + \mathcal{S_{G}}
  \label{enhanced shape prior}
\end{eqnarray}

In conclusion, original shape priors are enhanced with global and local characteristics \cite{su2022yolo}. Global shape priors can model the inter-class relations, which bear coarse shape priors with sufficient global texture information based on the self-attention block. Local shape priors show finer shape information via the introduction of inductive bias based on convolution. Besides, original skipped features are further enhanced via the interaction with global shape priors, which will promote generating features with discriminative shape representations and global contexts, then acquire more precise predicted masks.

% Transformer models including TransUNet, Swin-UNet, TransBTS, TransBTSV2, UNETR and Swin UNETR are trained with pretrained models.
\begin{table}[!t]
  \begin{center}
  \caption{Comparison with other models on BraTS 2020. (ET: the GD-enhancing tumor, WT: the whole tumor, TC: the tumor core, Mean: the average evaluation metric of three regions.)}
  \label{tab BraTS}
  \resizebox{\columnwidth}{!}{
  \begin{tabular}{ccccccccccc}  
  \hline  
  \multirow{2}*{Method} & \multicolumn{4}{c}{Dice score (\%) $\uparrow$} & \multicolumn{4}{c}{$HD_{95}$ (mm) $\downarrow$} & \multirow{2}*{Params} & \multirow{2}*{FLOPs} \\  
  \cmidrule(r){2-5}  \cmidrule(r){6-9}
  & ET & WT & TC & Mean & ET & WT & TC & Mean \\
  \hline
  3D UNet \cite{cciccek20163d}  & 77.85 & 90.41 & 83.26 & 83.84 & 17.94 & 4.90 & 5.77 & 9.53  & 16.47 &  516.71 \\
  Liu et al. \cite{liu2021brain}  & 76.37 & 88.23 & 80.12 & 81.57 & 21.39 & 6.68 & 6.49 & 11.52  & - & - \\
  Vu et al. \cite{vu2021multi}  & 77.17 & 90.55 & 82.67 & 83.46 & 27.04 & 4.99 & 8.63 & 13.55  & - & - \\
  Nguyen et al. \cite{nguyen2021enhancing}  & 78.43 & 89.99 & 84.22 & 84.21 & 24.02 & 5.68 & 9.57 & 13.09  & - & - \\
  ResUNet \cite{he2016deep} & 78.64 & 90.48 & \underline{85.18} & 84.77 & 17.77 & 6.56 & \underline{5.46} & 9.93  & 17.16  & 334.31  \\
  TransUNet \cite{chen2021transunet}  & 78.42 & 89.46 & 78.37 & 82.08 & 12.85 & 5.97 & 12.84 & 10.55  & 105.18  & 1035.52 \\
  Swin UNETR \cite{tang2022self}  & 79.61 & 89.51 & 84.69 & 84.60 & 14.61  & 11.18 & 6.10 & 10.63 & 62.19  & 790.61  \\
  UNeXt \cite{valanarasu2022unext}  & 76.49 & 88.70 & 81.37 & 82.19 & 16.61  & 4.98 & 11.50 & 11.04  &  4.02 & 14.92 \\
  TransBTS \cite{wang2021transbts}  & 78.73 & 90.09 & 81.73 & 83.52 & 17.95 & 4.96 & 9.77 & 10.89  & 32.99  & 412.97 \\
  TransBTSV2 \cite{li2022transbtsv2}  & \underline{79.63} & \underline{90.56} & 84.50 & \underline{84.90} & \underline{12.52} & \underline{4.27} & 5.56 & \underline{7.45}  &  15.30 &  320.54 \\
  Ours  & \textbf{79.70} & \textbf{91.08} & \textbf{85.35} & \textbf{85.38} & \textbf{12.06} & \textbf{3.92} & \textbf{5.08} & \textbf{7.02}  & 43.53  & 438.23  \\
  \hline  
  \end{tabular}}
  \end{center}
\end{table}

\section{Experiment}
\subsection{Datasets}
In this work, we conduct experiments on three public datasets for segmentation including the Brain Tumor Segmentation (BraTS) 2020 challenge \cite{menze2014multimodal, bakas2017advancing, bakas2018identifying}, the Large Scale Vertebrae Segmentation Challenge (VerSe 2019) \cite{sekuboyina2021verse} and the Automatic Cardiac Diagnosis Challenge (ACDC) \cite{bernard2018deep}.

% , loffler2020vertebral, liebl2021computed

\noindent\textbf{BraTS 2020:} This MRI dataset contains 369 training cases, 125 validation cases and 166 testing cases. Each case bears the same volume size $155 \times 240 \times 240$ and the same voxel space $1 \times 1 \times 1$ mm. Besides, each sample consists of four modality inputs, which are T1, T1-weighted, T2-weighted, and T2-FLAIR. The segmentation ground truth contains four classes, label 0 for background, label 1 for non-enhancing tumor core (NET), label 2 for peritumoral edema (ED), and label 4 for GD-enhancing tumor (ET). And the final evaluation metrics are Dice scores \cite{milletari2016v} and $95\%$ Hausdorff distance $HD_{95}$ \cite{huttenlocher1993comparing} on three regions, ET region (label 4), tumor core (TC, including label 1 and 4), the whole tumor (WT, containing label 1, 2 and 4). Furthermore, we introduce the average Dice score and $HD_{95}$ for an average evaluation of three regions.

\noindent\textbf{VerSe 2019:} This CT dataset is composed of 80 training cases, 40 validation cases, and 40 testing cases. There are 26 segmentation classes, including label 0 for the background and label $1\mbox{-}25$ for 25 vertebrae. Of all 25 vertebrae, label $1\mbox{-}7$ represents cervical vertebrae, label $8\mbox{-}19$ for thoracic vertebrae and label $20\mbox{-}25$ for lumbar vertebrae. Different samples show different field of views (FOVs), which means they may have different kinds of vertebrae. Here we select Dice scores and $HD_{95}$ for cervical, thoracic, and lumbar. Besides, mean and median values for all testing cases are also reported.

\noindent\textbf{ACDC:} This dataset involves 100 MRI scans from 100 patients. The target ROIs are the left ventricle (LV), right ventricle (RV), and myocardium (Myo). And we follow the data split setting of nnFormer \cite{zhou2021nnformer}, with 70 training cases, 10 validation cases, and 20 testing cases.

% Transformer models including TransUNet, CoTr and Swin UNETR are trained with pretrained models.
\begin{table}[!t]
  \begin{center}
  \caption{Comparison with other models on VerSe 2019. (Cerv: Cervical vertebrae, Thor: Thoracic vertebrae, Lumb: Lumbar vertebrae, Mean: the average evaluation metric of all vertebrae, Median: the median evaluation metric of all vertebrae.)}
  \label{tab verse}
  \resizebox{\columnwidth}{!}{
  \begin{tabular}{ccccccccccccc}
  \hline  
  \multirow{2}*{Method} & \multicolumn{5}{c}{Dice score (\%) $\uparrow$} & \multicolumn{5}{c}{$HD_{95}$ (mm) $\downarrow$}  & \multirow{2}*{Params} & \multirow{2}*{FLOPs} \\  
  \cmidrule(r){2-6}  \cmidrule(r){7-11}
  & Cervical & Thoracic & Lumbar & Mean & Median & Cervical & Thoracic & Lumbar & Mean & Median \\
  \hline  
  3D UNet \cite{cciccek20163d}  & 83.10 & 78.37 & 70.88 & 81.28 & 87.54 & 3.26 & 6.27 & 8.50 & 5.80 & 4.02 & 16.49 & 520.98 \\
  VNet \cite{milletari2016v}  & 86.32 & 87.78 & 73.45 & 85.57 & 92.14 & 2.19 & 3.37 & 8.48 & 4.11 & \underline{1.85} & 45.73 & 957.88 \\
  nnUNet \cite{isensee2021nnu}  & 87.81 & \textbf{88.80} & \textbf{74.96} & \underline{86.59} & \underline{92.62} & 2.52 & 3.04 & \textbf{7.10} & 4.09 & 2.11 & 30.90  & 502.80 \\
  TransUNet (3D) \cite{chen2021transunet}  & 85.49 & 82.67 & 73.88 & 83.53 & 88.06 & 2.02 & 3.73 & 7.89 & 4.16 & 2.51 &  146.68 & 682.96 \\
  CoTr \cite{xie2021cotr}  & 81.48 & 79.68 & 68.83 & 80.59 & 85.72 & 3.92 & 9.88 & 14.34 & 9.05 & 6.34 & 48.53 &  507.68 \\
  UNeXt \cite{valanarasu2022unext}  & 77.00 & 86.73 & 71.06 & 83.36 & 88.39  & 3.44 & \underline{2.97} & 9.47 & 4.43 & 2.63  & 4.02 & 12.17  \\
  maskformer \cite{cheng2021per}  & 76.22 & 80.87 & 72.01 & 83.24 & 89.92  & 2.32 & 7.47 & 9.08 & 6.29 & 2.98  & 64.40 & 943.48 \\
  EG-Trans3DUNet \cite{you2022eg}  & 83.67 & 82.41 & 74.11 & 86.01 & 91.12 & 2.37 & 4.46 & 8.12 & \underline{4.03} & 2.54  & 161.89 & 748.20 \\
  Swin UNETR \cite{tang2022self}  & \underline{89.30} & 81.43 & 73.36 & 83.46 & 88.91  & \underline{1.85} & 5.90 & 8.81 & 5.75 & 3.95  & 62.19  & 732.18 \\
  Ours  & \textbf{89.87} & \underline{88.69} & \underline{74.15} & \textbf{87.65} & \textbf{93.43} & \textbf{1.70} & \textbf{2.49} & \underline{7.72} & \textbf{3.58} & \textbf{1.78}  & 46.55  & 457.63 \\
  \hline  
  \end{tabular}}
  \end{center}
\end{table}

% Transformer models including TransUNet, Swin-UNet and UNETR are trained with pretrained models.
\begin{table}[!t]
  \begin{center}
  \caption{Comparison with other models on ACDC. Metric: Dice scores (\%). (RV: right ventricle, Myo: myocardium, LV: left ventricle, Mean: the average evaluation metric of all regions.)}
  \label{tab ACDC}
    \setlength{\tabcolsep}{1.0mm}{
  \begin{tabular}{ccccccc}  
  \hline  
  Method & RV & Myo & LV & Mean & Params (M) & FLOPs (G)  \\  
  \hline  
  R50 Att-UNet \cite{oktay2018attention} & 87.58 & 79.20 & 93.47 & 86.75 & 107.60 & 639.09 \\
  TransUNet \cite{chen2021transunet}  & 88.86 & 84.54 & 95.73 & 89.71 & 105.50  & 643.20 \\
  Swin-UNet \cite{cao2021swin}  & 88.55 & 85.62 & 95.83 & 90.00 & 27.17 & 118.00 \\
  UNETR \cite{hatamizadeh2022unetr}  & 85.29 & 86.52 & 94.02 & 88.61 & 93.10 & 195.62 \\
  MISSFormer \cite{huang2021missformer}  & 86.36 & 85.75 & 91.59 & 87.90 & 42.46 & 187.80 \\
  % LeViT-UNet-384 \cite{graham2021levit}  & 89.55 & 87.64 & 93.76 & 90.32  \\
  Maskformer \cite{cheng2021per}  & 87.89 & 87.34 & 94.92 & 90.05 & 45.87 & 712.69 \\
  nnUNet \cite{isensee2021nnu}  & 90.24 & \underline{89.24} & 95.36 & 91.61 & 30.43  & 471.86\\
  nnFormer \cite{zhou2021nnformer}  & \underline{90.94} & \textbf{89.58} & \underline{95.65} & \underline{92.06} & 147.11 & 291.76 \\
  Ours  & \textbf{92.28} & 88.13 & \textbf{96.61} & \textbf{92.34} & 41.86 & 257.90 \\
  \hline  
  \end{tabular}}
  \end{center}
\end{table}

\begin{comment}
\begin{figure*}[ht]
\centerline{\includegraphics[scale=0.14]{ape16-fix.png}}
\centerline{\includegraphics[scale=0.14]{dpe54-fix.png}}
\caption{Overview of Network Architecture. (a) The whole network architecture of EG-Trans3DUNet. (b) Edge detection block
(c) Network structure for Local and Global Transformer Encoder.
(d) Global information extraction block}
\end{figure*}
\end{comment}

\begin{comment}
\begin{figure}
  \centering
  \includegraphics[scale=0.5]{ape16-fix.png}
  \hspace{1in}
  \includegraphics[scale=0.5]{dpe54-fix.png}
  \caption{ape and dpe}
\end{figure}
\end{comment}

\begin{comment}
%% caption放在小页里边
\begin{figure}
  \begin{minipage}[t]{0.5\linewidth}
    \centering
    \includegraphics[scale=0.4]{ape16-fix-r.png}
    \caption{Absolute Positional Embeddings}
    \label{fig:side:a}
  \end{minipage}%
  \begin{minipage}[t]{0.5\linewidth}
    \centering
    \includegraphics[scale=0.4]{dpe54-fix-r.png}
    \caption{Multi-scale Dynamic Positional Embeddings}
    \label{fig:side:b}
  \end{minipage}
\end{figure}
\end{comment}

\begin{figure*}[!t]
\centerline{\includegraphics[width=0.85\linewidth]{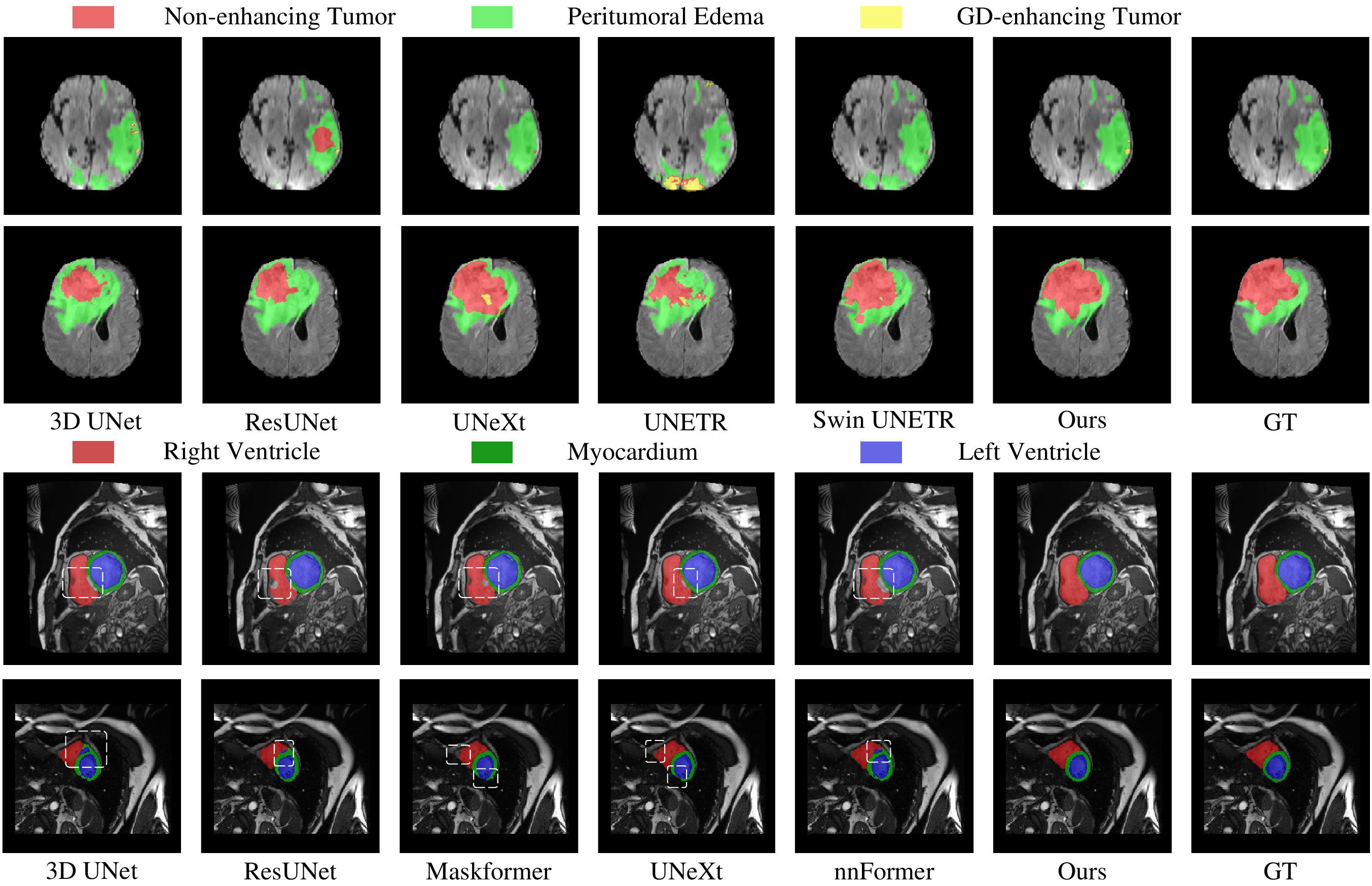}}
\caption{Predicted masks of different models on BraTS 2020 and ACDC. Each column refers to a segmentation result of a network model.}
\label{visualizations brats and ACDC}
\end{figure*}

%For the design of our proposed network as shown in Figure \ref{segmentation paradigm}, we adopt the classic U-shape network. Specifically, we employ ResNet-50 \cite{he2016deep} as the encoder. And the basic component of decoder is the residual block similar to Res-block in \cite{tang2022self}.

%  All three public datasets are preprocessed with the z-score normalization \cite{isensee2021nnu}.
\noindent \textbf{Implementation Details.} The proposed model is implemented with PyTorch 1.8.0 and trained on 2 NVIDIA Telsa V100, with a batch size of 2 in each GPU. For the BraTS 2020 dataset, all models are trained with the AdamW \cite{loshchilov2017decoupled} optimizer for 2000 epochs, with a warm-up cosine scheduler for the first 50 epochs. The initial learning rate is set as $8e\mbox{-}4$ with $1e\mbox{-}5$ weight decay. And the size of cropped patches is $128 \times 128 \times 128$. We do not utilize complicated data augmentations like previous works \cite{isensee2021nnu, zhou2021nnformer}. Instead, we adopt strategies of random mirror flipping, random rotation, random intensity shift and scale. For the VerSe 2019 dataset, we train all models for 1000 epochs. All preprocessed cases are cropped with a patch size of $128 \times 160 \times 96$. Random rotation between [$-15^{\circ}, 15^{\circ}$] and random flipping along the XOZ or YOZ plane are employed for the data diversity. For the ACDC dataset, models are trained for 1500 epochs and the patch size is set as $20 \times 256 \times 256$. Similarly, we augment the cardiac data with random rotation and random flipping. And for both VerSe 2019 and ACDC, we choose the Adamw optimizer with the initial learning rate set as $5e\mbox{-}4$ and the cosine warm-up strategy for 50 epochs during training. Following the setup in \cite{isensee2021nnu}, we choose a sum of Dice loss and cross-entropy loss for model training.

\begin{figure}[!t]
\centerline{\includegraphics[width=0.9\columnwidth]{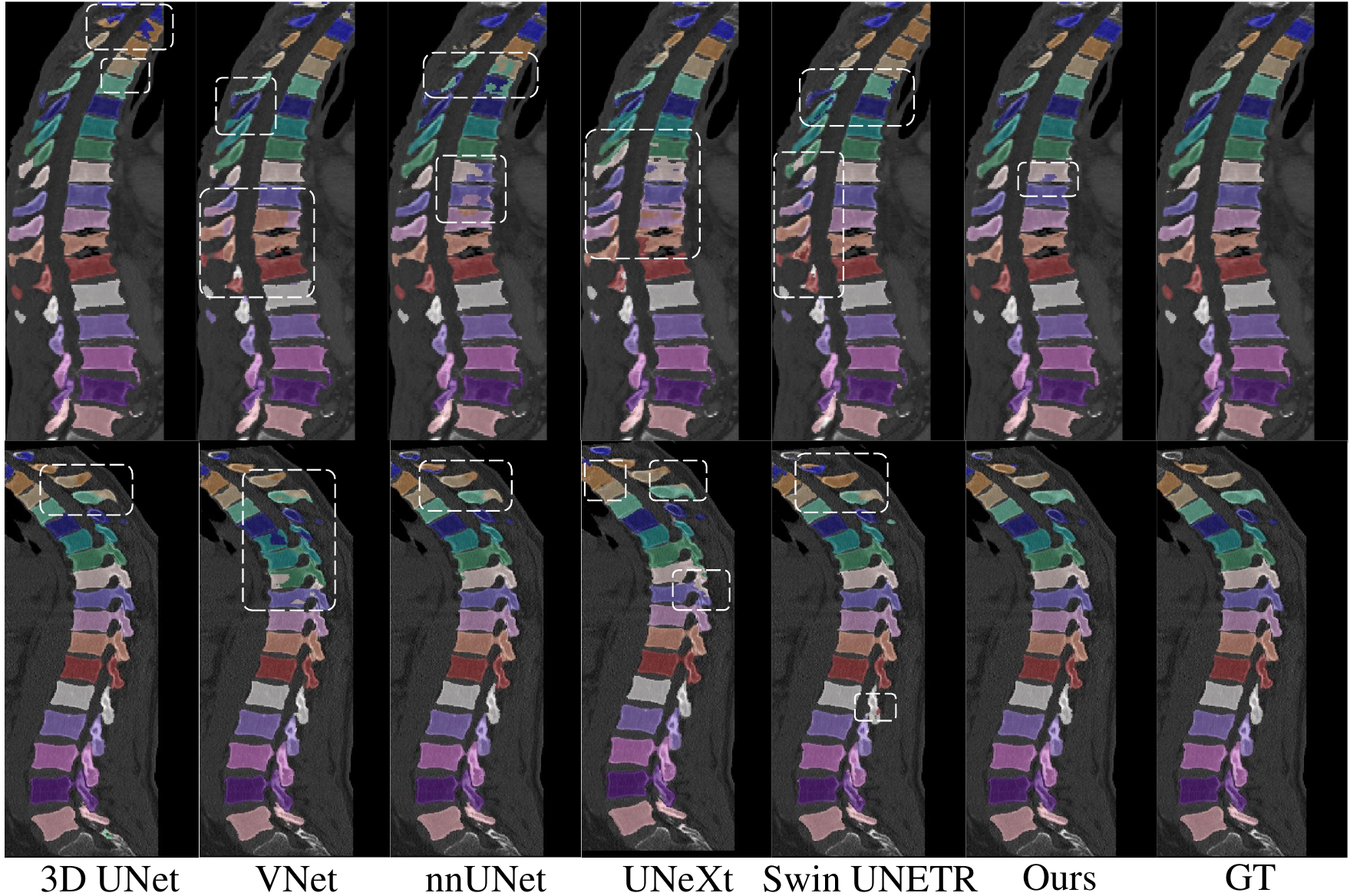}}
\caption{Predicted masks of different models on VerSe 2019.}
\label{visualizations verse}
\end{figure}
\subsection{Experimental Results}
\textbf{Brain tumor segmentation:} Table \ref{tab BraTS} illustrates the quantitative segmentation performance of our proposed model compared with other CNNs and Transformer-based models. It can be figured out that our model shows absolute superiority on the Dice score and $HD_{95}$ of all three regions. Compared with TransBTSV2 \cite{li2022transbtsv2}, our model achieves higher Dice scores of $0.07\%$, $0.52\%$, $0.85\%$ and a lower $HD_{95}$ of $0.46mm$, $0.35mm$, $0.48mm$ on the ET, WT and TC region. This improvement results from the fact that enhanced shape priors serve as anatomical priors to be injected into networks, which ease the dependence on the final learnable prototype. It is worth mentioning that there is a significant improvement in the metrics for Region WT and TC. We claim that our model with refined shape priors is aimed at enhancing shape representations for region WT and TC because they bear a relatively fixed shape in contrast to region ET. We will prove this viewpoint in the ablation study on the generalization abilities of SPM with different network structures.

\begin{figure*}[!t]
  \setlength{\abovecaptionskip}{-1pt}
\centerline{\includegraphics[width=0.9\linewidth]{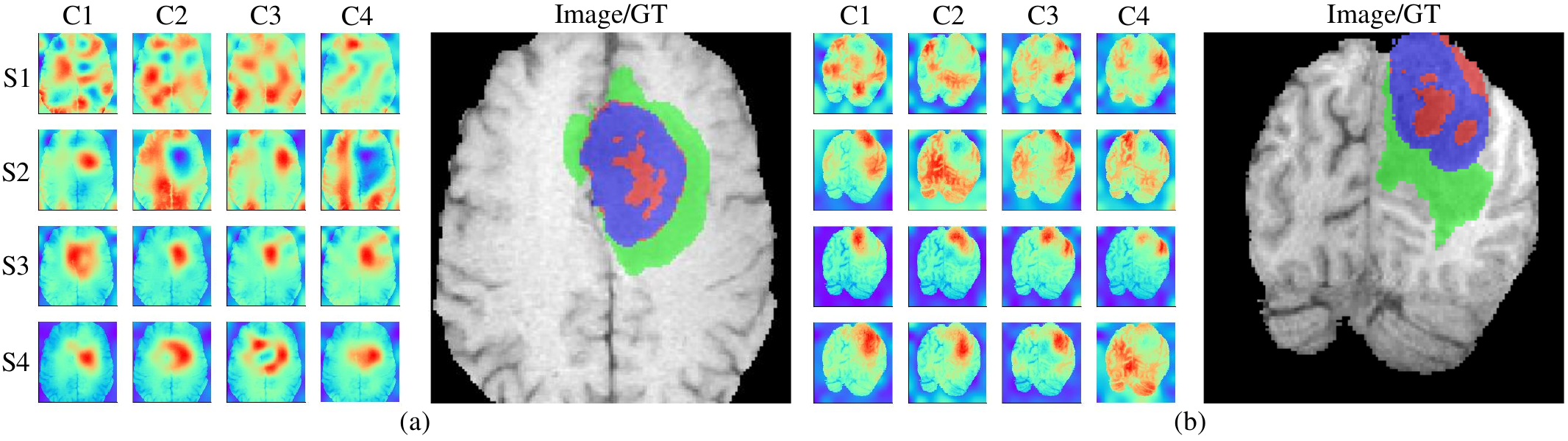}}
\caption{Different channels (from Channel 1 to Channel 4) and different stages (from Stage 1 to Stage 4) of generated shape priors on two cases of BraTS 2020. $C_{i}$ and $S_{i}$ ($i$=0, 1, 2, 3) represent the respective channels and stages.}
\label{shape priors}
\end{figure*}

\noindent\textbf{Vertebrae segmentation:} To further evaluate the performance of our proposed model, we conduct experiments on VerSe 2019. Table \ref{tab verse} presents the segmentation performance on the hidden test dataset. Our model outperforms the powerful nnUNet on the metric of cervical, and thoracic vertebrae. Specifically, there is $2.06\%$, $1.06\%$ Dice score increases and $0.82$mm, $0.33$mm $HD_{95}$ decreases for cervical and the whole spine. Besides, different cases have different field-of-views (FoVs), which brings difficulty for models to identify the last several vertebrae. And nnUNet is superior to other models including ours on the ability to localize and segment lumbar vertebrae (Label $20\mbox{-}25$). Furthermore, in contrast with other models, we achieve the highest median Dice score $93.43\%$ and $1.78$mm, which means that our model can boost the overall performance of testing cases. Figure \ref{visualizations verse} illustrates that our model presents better visualization results, with more consistent predictions in a singular vertebra. This phenomenon indicates that learnable explicit shape priors for vertebrae are truly effective in the refinement of predicted masks.

\begin{figure}[!t]
\centerline{\includegraphics[width=\columnwidth]{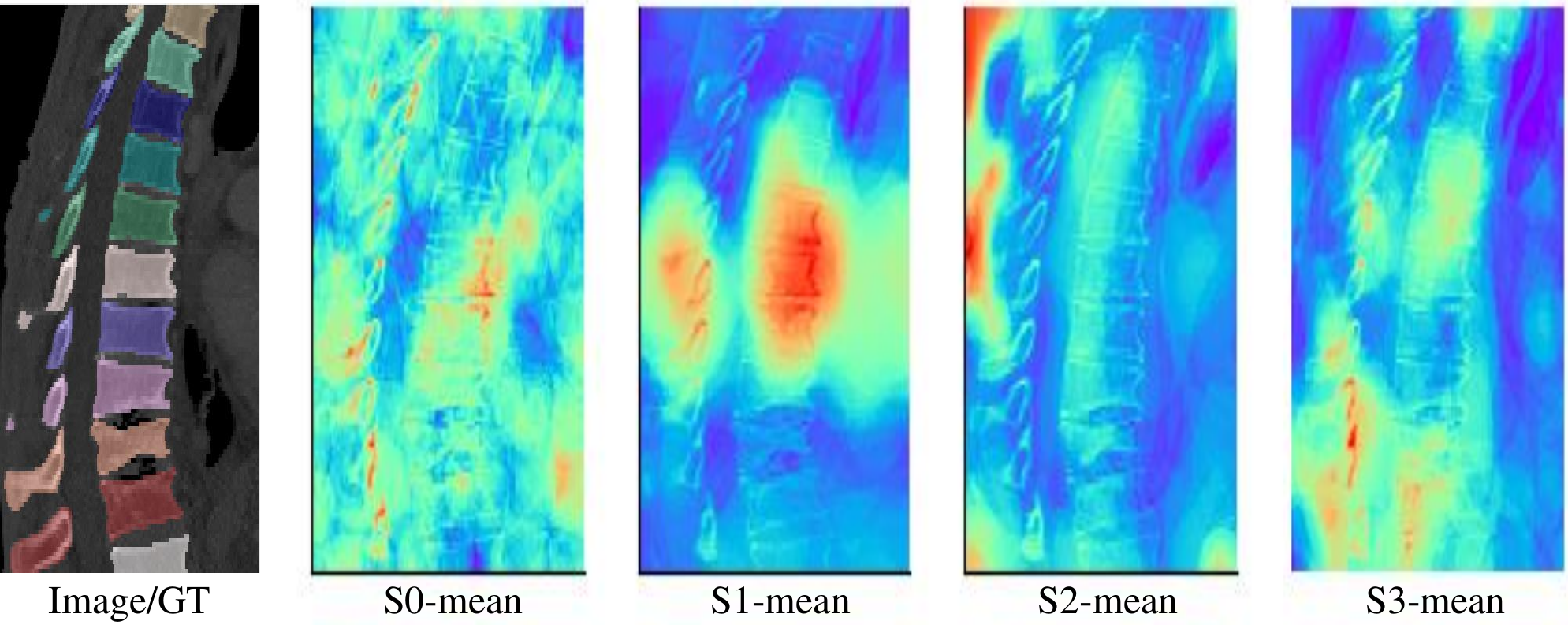}}
%\setlength{\abovecaptionskip}{1pt}
%\vspace{-0.8cm}
  \setlength{\abovecaptionskip}{-1pt}
\caption{Mean shape priors on Channel 1 to Channel 26 of different stages (from Stage 0 to Stage 3) on two cases of VerSe 2019. $S_{i}$-mean ($i$=0, 1, 2, 3) means average shape priors on channels of the $i_{th}$ stage.}
\label{verse shape priors}
\end{figure}

\begin{figure}[!t]
\centerline{\includegraphics[width=\columnwidth]{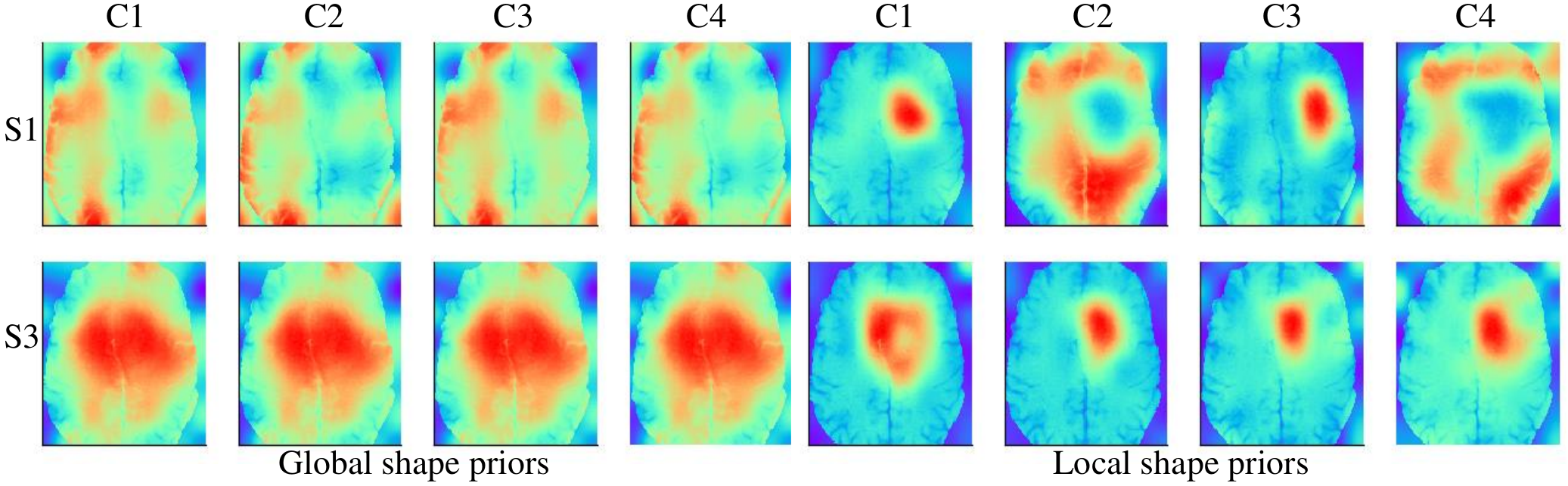}}
  \setlength{\abovecaptionskip}{-1pt}
\caption{Different channels and different stages of global and local shape priors on case (a) in Figure \ref{shape priors}. $C_{i}$ and $S_{i}$ ($i$=0, 1, 2, 3) represent the respective channels and stages.}
\label{global and local shape priors}
\end{figure}

\noindent\textbf{Automated cardiac segmentation:} We also conduct quantitative and qualitative experiments on the ACDC dataset. As shown in Table \ref{tab ACDC}, our model outperforms the previous SOTA model nnFormer \cite{zhou2021nnformer} on the evaluation metrics for the RV and LV regions. Specifically, the Dice scores of RV and LV reach $92.28\%$ and $96.61\%$, with $1.34\%$ and $0.96\%$ higher than that of nnFormer. Figure \ref{visualizations brats and ACDC} demonstrates that our model outputs more accurate segmentation masks, particularly in the RV and LV regions. However, the segmentation performance for Myo is lower than that of nnUNet \cite{isensee2021nnu} and nnFormer. We argue that networks will be more focused on larger regions and ignore smaller regions due to the label imbalance between RV, LV and Myo \cite{ma2021loss}. Besides, this MRI dataset shows a large voxel space, which will aggravate the effect of label imbalance. And the unique and effective resampling strategy of nnUNet and nnFormer will improve the imbalanced distribution of the myocardium tissue, which results in stronger attention of models on this region. Thus, nnUNet and nnFormer achieve a higher Dice score on the Myo region.

%\vspace{-0.8cm}  %调整图片与上文的垂直距离

%\setlength{\abovecaptionskip}{-0.2cm}   %调整图片标题与图距离

%\setlength{\belowcaptionskip}{-1cm}   %调整图片标题与下文距离

\begin{comment}

% \includegraphics[scale=0.50]
\begin{figure}[!t]
\centerline{\includegraphics[width=\columnwidth]{feature_enhance_acdc.pdf}}
\caption{Qualitative comparisons of baseline and baseline + SPM on BraTS 2020, VerSe 2019 and ACDC.}
\label{feature enhance: acdc}
\end{figure}
    
\end{comment}

% \includegraphics[scale=0.46]
\begin{figure}[!t]
  \setlength{\abovecaptionskip}{-1pt}
\centerline{\includegraphics[width=\columnwidth]{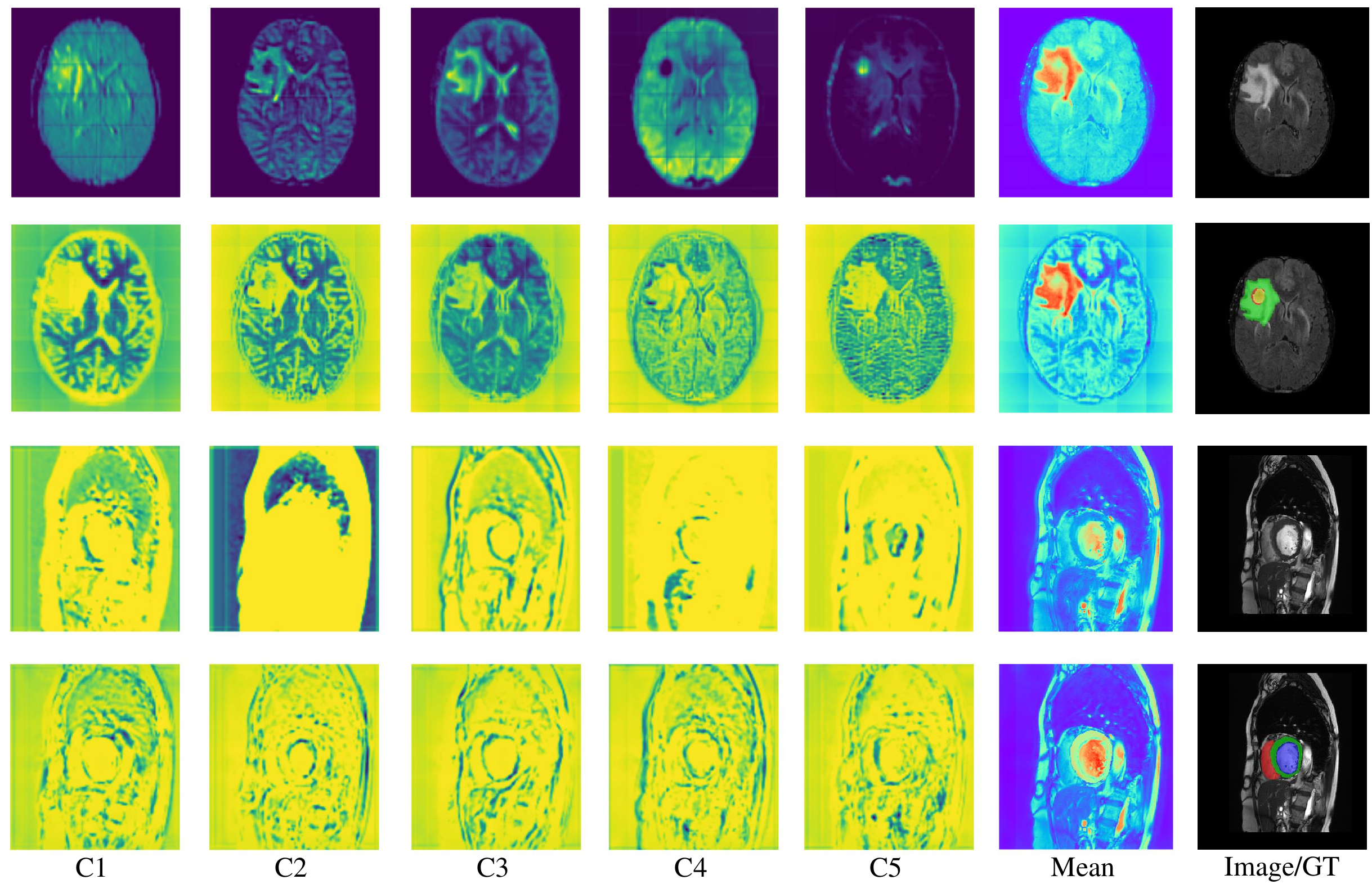}}
\caption{Feature visualizations of baseline and baseline with SPM on BraTS 2020 and ACDC. $C_{i}$ ($i$=1, 2, 3, 4, 5) represents the respective channel. 'Mean' refers to mean features across all 32 channels.}
\label{feature enhance: brats and acdc}
\end{figure}

\subsection{Visualizations of shape priors and skipped features}\label{visualization section}
In this section, we probe into the qualitative results of shape priors and skipped features. In fact, they are mutually enhanced. We first discuss the impact of skipped features on explicit shape priors. As mentioned in section \ref{different paradigms about shape priors}, explicit shape priors are iteratively updated under the guidance of skipped convolutional features, then optimized shape priors will activate regions of interest. We visualize two cases from the BraTS 2020 dataset in Figure \ref{shape priors}. Case (a) illustrates generated explicit shape priors from different stages. Specifically, shape priors consist of N-channel attention maps, in which N refers to the number of segmentation classes, and each row represents shape priors from each stage. We can figure it out that shape priors reveal more accurate activation maps for the ground truth region as the top-to-down process. In particular, wrongly activated regions in the first stage will be suppressed in the second and third stages of SPM. Here in our visualization results, there exists a phenomenon called the reverse activation \cite{fan2020pranet}, which means that all regions except the ground truth area are activated. A canonical example is visualized in the last stage and last channel of shape priors in case (b). We claim that this phenomenon results from the global shape priors, which bring global contexts and sufficient texture information for the whole region, even including regions from the background. In essence, it is simple to locate the ROIs via reverse attention, in which ROIs are highlighted with distinct contours. From this point of view, reverse activation is similar to positive activation.

\begin{table}[!t]
  \begin{center}
  \caption{Ablation study about the plug-and-play characteristic and structural composition of SPM on the BraTS 2020, VerSe 2019 and ACDC dataset.}
  \label{ablation}
  \resizebox{\columnwidth}{!}{
  \begin{tabular}{ccccccccc}  
  \hline  
  \multicolumn{9}{c}{\textbf{BraTS}} \\
  \hline
  \multirow{2}*{Method} & \multicolumn{4}{c}{Dice score (\%) $\uparrow$} & \multicolumn{4}{c}{$HD_{95}$ (mm) $\downarrow$}  \\  
  \cmidrule(r){2-5}  \cmidrule(r){6-9}
  & ET & WT & TC & Mean & ET & WT & TC & Mean \\
  \hline
  3D UNet \cite{cciccek20163d}  & 82.38 & 91.89 & 90.24 & 88.16 & 4.20 & 4.93 & 3.35 & 4.16 \\
  + SPM  & \textbf{82.86} & \textbf{92.42} & \textbf{90.46} & \textbf{88.58} & \textbf{3.77} & \textbf{4.26} & \textbf{3.28} & \textbf{3.77} \\
    \hline
  ResUNet \cite{he2016deep} & 83.33 & 92.36 & 89.47 & 88.38 & 4.09 & 6.38 & 4.09 & 4.85 \\
  + SPM & \textbf{84.26} & \textbf{92.70} & \textbf{91.05} & \textbf{89.34} & \textbf{4.04} & \textbf{3.30} & \textbf{3.56} & \textbf{3.63} \\
    \hline
  UNETR \cite{hatamizadeh2022unetr}  & 79.84 & 88.81 & 83.22 & 83.96 & 8.95 & 17.99 & 14.02 & 13.65\\
  + SPM  & \textbf{80.25} & \textbf{90.08} & \textbf{85.69} & \textbf{85.34} & \textbf{8.41}  & \textbf{15.16} & \textbf{10.65} & \textbf{11.41}  \\
    \hline
  UNeXt \cite{valanarasu2022unext}  & 79.72 & 91.61 & 89.32 & 86.88 & 4.06  & 4.64 & 4.46 & 4.39 \\
  + SPM  & \textbf{80.61} & \textbf{91.89} & \textbf{89.75} & \textbf{87.42} & \textbf{3.40} & \textbf{4.11} & \textbf{4.39} & \textbf{3.97} \\
    \hline
  Swin UNETR \cite{tang2022self}  & 83.51 & 91.95 & 90.20 & 88.55 & 5.95 & 8.99 & \textbf{4.93} & 6.63 \\
  + SPM  & \textbf{83.68} & \textbf{92.92} & \textbf{90.36} & \textbf{88.99} & \textbf{4.83} & \textbf{6.73} & 6.39 & \textbf{5.98} \\
  \hline  

  Baseline & 83.33 & 92.36 & 89.47 & 88.38 & 4.09 & 6.38 & 4.09 & 4.85 \\
  + CUB  & 84.19 & 92.47 & 90.40 & 89.02 & \textbf{3.86} & 3.52 & 4.09 & 3.82 \\
  + SUB \& CUB & \textbf{84.26} & \textbf{92.70} & \textbf{91.05} & \textbf{89.34} & 4.04 & \textbf{3.30} & \textbf{3.56} & \textbf{3.63} \\

  \hline 
  \end{tabular}}

\resizebox{\columnwidth}{!}{
  \begin{tabular}{ccccccccc}  
  \hline 
    \multicolumn{9}{c}{\textbf{VerSe 2019}} \\
    \hline
  \multirow{2}*{Method} & \multicolumn{4}{c}{Dice score (\%) $\uparrow$} & \multicolumn{4}{c}{$HD_{95}$ (mm) $\downarrow$}  \\  
  \cmidrule(r){2-5}  \cmidrule(r){6-9}
  & Cerv & Thor & Lumb & Mean & Cerv & Thor & Lumb & Mean \\
  \hline  
  3D UNet \cite{cciccek20163d}  & 83.10 & 78.37 & 70.88 & 81.28 & 3.26 & 6.27 & \textbf{8.50} & 5.80  \\
  + SPM  & \textbf{85.30} & \textbf{84.79} & \textbf{72.28} & \textbf{84.16} & \textbf{2.44} & \textbf{5.11} & 9.02 & \textbf{5.39} \\
    \hline
  ResUNet \cite{he2016deep}  & 88.59 & 84.50 & 73.08 & 84.98 & 2.18 & 4.16 & 9.42 & 4.43 \\
  + SPM  & \textbf{89.87} & \textbf{88.69} & \textbf{74.15} & \textbf{87.65} & \textbf{1.70} & \textbf{2.49} & \textbf{7.72} & \textbf{3.58} \\

    \hline
  UNETR \cite{hatamizadeh2022unetr}  & 72.10 & 69.33 & \textbf{67.07} & 73.86 & 6.22 & 9.13 & 13.07 & 9.27 \\
  + SPM  & \textbf{79.95} & \textbf{74.62} & 66.32 & \textbf{75.86}  & \textbf{3.97} & \textbf{6.89} & \textbf{12.19} & \textbf{7.81} \\
    \hline
  UNeXt \cite{valanarasu2022unext}  & 77.00 & \textbf{86.73} & 71.06 & 83.36 & 3.44 & \textbf{2.97} & 9.47 & 4.43 \\
  + SPM  & \textbf{85.35} & 86.18 & \textbf{73.15} & \textbf{84.62} & \textbf{2.12} & 3.29 & \textbf{7.95} & \textbf{4.06} \\
    \hline
  Swin UNETR \cite{tang2022self}  & \textbf{89.30} & 81.43 & 73.36 & 83.46 & \textbf{1.85} & 5.90 & \textbf{8.81} & 5.75 \\
  + SPM  & 81.82 & \textbf{85.58} & \textbf{73.61} & \textbf{84.69} & 3.30 & \textbf{3.61} & 8.83 & \textbf{4.70} \\
  \hline  

  % Baseline  & 88.97 & 84.94 & 73.44 & 85.87 & 91.13 & 2.19 & 4.05 & 8.10 & 4.13 & 2.72 \\
  Baseline  & 88.59 & 84.50 & 73.08 & 84.98 & 2.18 & 4.16 & 9.42 & 4.43 \\
  + CUB  & 82.48 & 85.63 & 74.02 & 84.46 & 3.27 & 3.92 & 8.61 & 5.03 \\
  + SUB \& CUB  & \textbf{89.87} & \textbf{88.69} & \textbf{74.15} & \textbf{87.65} & \textbf{1.57} & \textbf{2.49} & \textbf{7.72} & \textbf{3.58} \\
    \hline
  
  \end{tabular}}

%\resizebox{\columnwidth}{!}{
\resizebox{\columnwidth}{!}{
  \begin{tabular}{ccccccccc}  
  \hline  
      \multicolumn{9}{c}{\textbf{ACDC}} \\
      \hline
  \multirow{2}*{Method} & \multicolumn{4}{c}{Dice score (\%) $\uparrow$} & \multicolumn{4}{c}{$HD_{95}$ (mm) $\downarrow$}  \\  
  \cmidrule(r){2-5}  \cmidrule(r){6-9}
  & RV & Myo & LV & Mean & RV & Myo & LV & Mean \\
  \hline  
  3D UNet \cite{cciccek20163d} & 91.13 & 85.75 & 95.95 & 90.94 & 1.46 & 1.17 & \textbf{1.07} & 1.24 \\
  + SPM & \textbf{91.78} & \textbf{86.73} & \textbf{96.29} & \textbf{91.60} & \textbf{1.45} & \textbf{1.12} & 1.09 & \textbf{1.22} \\
    \hline
  ResUNet \cite{he2016deep}  & 91.66 & 85.26 & 95.72 & 90.88 & 1.47 & 1.21 & \textbf{1.09} & 1.26 \\
  + SPM  & \textbf{92.28} & \textbf{88.13} & \textbf{96.61} & \textbf{92.34} & \textbf{1.43} & \textbf{1.10} & \textbf{1.09} & \textbf{1.21} \\
    \hline
  UNeXt \cite{valanarasu2022unext}  & 91.85 & 86.91 & 96.34 & 91.70 & 1.47 & \textbf{1.09} & \textbf{1.05} & 1.21 \\
  + SPM  & \textbf{92.14} & \textbf{87.48} & \textbf{96.58} & \textbf{92.07} & \textbf{1.37} & 1.13 & \textbf{1.05} & \textbf{1.18} \\
    \hline
 
  Baseline  & 91.66 & 85.26 & 95.72 & 90.88 & 1.47 & 1.21 & \textbf{1.09} & 1.26 \\
  + CUB  & 91.60 & 87.69 & \textbf{96.61} & 91.97 & 1.46 & 1.15 & 1.10 & 1.24 \\
  + SUB \& CUB  & \textbf{92.28} & \textbf{88.13} & \textbf{96.61} & \textbf{92.34} & \textbf{1.43} & \textbf{1.10} & \textbf{1.09} & \textbf{1.21} \\
    \hline
 
  \end{tabular}}
  
  \end{center}
\end{table}

Further, we decompose shape priors into two components, global and local shape priors generated from SUB and CUB respectively. We visualize these two components of case (a) in Figure \ref{global and local shape priors}. Due to the self-attention module \cite{dosovitskiyimage}, global shape priors bear globalized receptive fields, containing contexts and textures. However, the structure of SUB lacks inductive bias to model local visual structures. Here we can discover that global shape priors are responsible for a coarse localization for the region of ground truths. And local shape priors generated from CUB can provide finer shape information for the ROIs via the introduction of convolutional kernels, which bear the inductive bias of locality. 

We then thoroughly analyze the impact of shape priors on skipped features with the direct comparison between original skipped features $F_{o}$ and enhanced skipped features $F_{e}$. In detail, specific channels of $F_{o}$ and $F_{e}$ are selected from all 32 channels for a qualitative visualization. As shown in Figure \ref{feature enhance: brats and acdc}, features in the tumor region are enhanced and some voxels which are not activated before, are highlighted after processing by SPM. Besides, via the introduction of global shape priors, skipped features are enriched with sufficient texture information for the whole region. We also explain the process of feature refinement with a cardiac CT case as shown in Figure \ref{feature enhance: brats and acdc}. The channel-average skipped features are refined with more attention on the LV and Myo regions.

\subsection{Ablation Studies and Discussions}
\subsubsection{Plug-and-play}
To prove the plug-and-play characteristic of SPM, we detailedly carry out ablation studies on the generalization ability of SPM on different network structures. Here we choose CNNs, Transformer-based and MLP-based models, including 3D UNet \cite{cciccek20163d}, ResUNet \cite{he2016deep}, UNETR \cite{hatamizadeh2022unetr}, Swin UNETR \cite{tang2022self} and UNeXt \cite{valanarasu2022unext}. For evaluations on the BraTS 2020 dataset, we report the segmentation performance on 41 split validation cases during training. For the ACDC dataset, each MRI scan consists of thick slices, which is not applicable to the input size of UNETR and Swin UNETR. Thus, we only report quantitative results on 3D UNet, ResUNet and UNeXt.

\begin{comment}
\begin{figure}[!t]
\centerline{\includegraphics[width=\columnwidth]{visualization for spm.pdf}}
\caption{Qualitative comparisons of baseline and baseline + SPM on BraTS 2020, VerSe 2019 and ACDC.}
\label{ablation visualizations}
\end{figure}    
\end{comment}

\begin{figure}[!t]
\centerline{\includegraphics[width=\columnwidth]{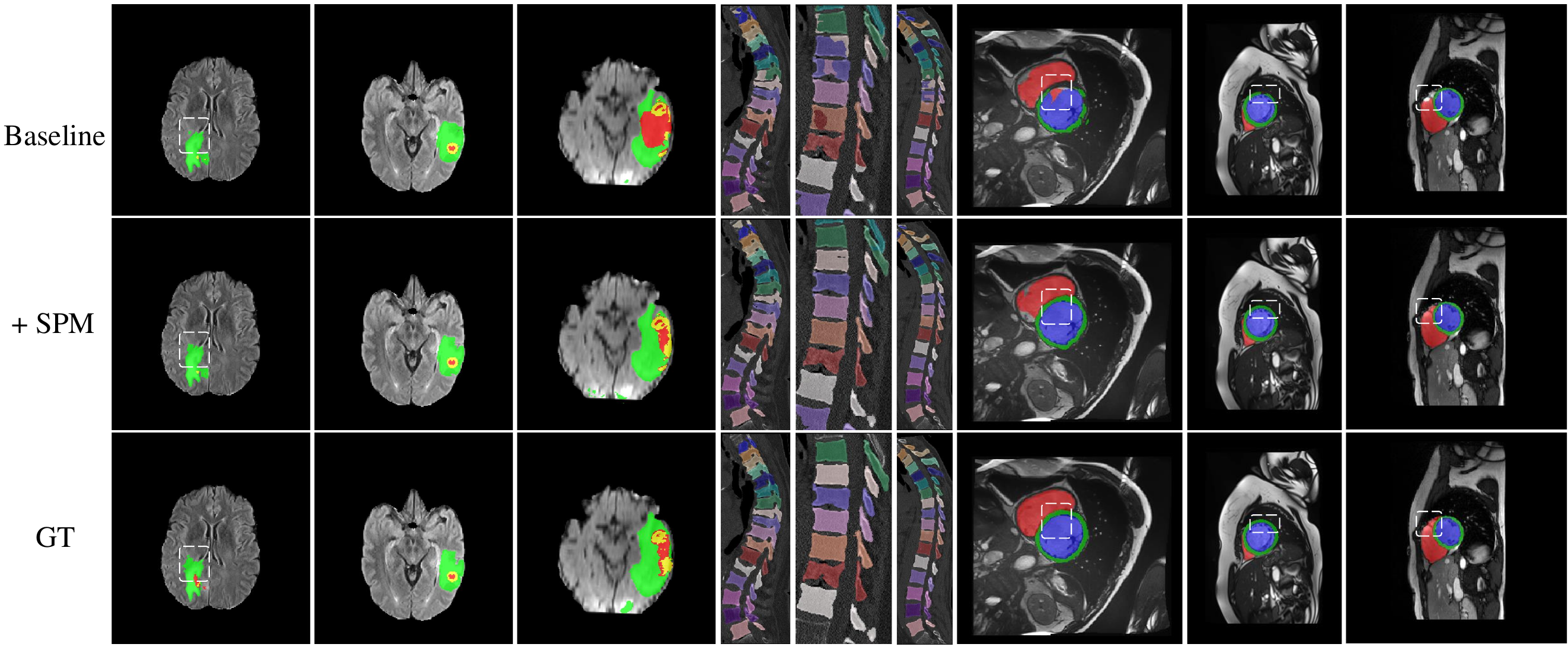}}
\caption{Qualitative comparisons of baseline and baseline with SPM on BraTS 2020, VerSe 2019 and ACDC.}
\label{ablation visualizations}
\end{figure}

According to Table \ref{ablation}, it can be observed that SPM can boost the segmentation performance of different networks. Specifically, SPM can bring an increase on ResUNet \cite{he2016deep}, with a Dice score increase of $0.77\%$, $0.51\%$, $1.48\%$ on the ET, WT, TC region. And combined with SPM, 3D UNet \cite{cciccek20163d}shows an improvement on the metric of $HD_{95}$, with a decrease of $0.43mm$, $0.67mm$, $0.07mm$ on region ET, WT, TC. Furthermore, SPM also upgrades the segmentation performance of Transformer-based models. SPM brings a Dice score increase of $1.27\%$ and $0.97\%$ on the WT region for UNETR \cite{hatamizadeh2022unetr} and Swin UNETR \cite{tang2022self}, which reveals the potential of SPM to significantly enhance the representation ability for regions with relatively regular shapes. And we can also explain this phenomenon from the perspective of inductive bias. With the introduction of shape priors, we introduce a strong inductive bias to Transformer-based models, which will relieve the requirements for a huge amount of datasets and accelerate the convergence of Transformers. Besides, SPM can improve the segmentation performance of the enhanced tumors, which bear various and irregular shapes. This phenomenon can be explained by the fact that global shape priors inject global texture information to skipped features as shown in Figure \ref{feature enhance: brats and acdc}, and the context information is effective to improve models' representation abilities for enhanced tumors.

As shown in Table \ref{ablation},  a significant improvement is obtained on the segmentation performance of the ACDC dataset after employing the proposed SPM to the baseline model. Particularly, the Dice score for the myocardium region increases by $0.98\%$, $2.87\%$, $0.57\%$ on 3D UNet \cite{cciccek20163d}, ResUNet \cite{he2016deep} and UNeXt \cite{valanarasu2022unext} respectively. We carry out a visualization comparison between ResUNet and ResUNet with SPM. Figure \ref{ablation visualizations} reveals that SPM can refine segmentation masks with a roughly circular shape, which benefits from the learnable shape priors. Besides, false positive predictions of RV introduce inconsistency to the segmentation result of LV, which will be suppressed by anatomical shape priors learned from training datasets.

Furthermore, we conduct comprehensive experiments on VerSe 2019 to evaluate the efficacy of SPM when plugged into CNNs and Transformer-based structures. As shown in Table \ref{ablation}, SPM brings considerable improvements on the Dice score and $95\%$ Hausdorff distance of cervical, thoracic and lumbar vertebrae. For ResUNet, the introduction of shape priors brings remarkable gains on the segmentation performance of cervical, thoracic, lumbar vertebrae, with $1.28\%$, $4.19\%$, $1.07\%$ Dice score increases and $0.48$mm, $1.67$mm, $1.70$mm $HD_{95}$ decreases. And for Swin UNETR \cite{tang2022self}, a Transformer-based model, substantial improvements have been achieved on the Dice score and $HD_{95}$ of thoracic vertebrae ($\uparrow4.15\%$, $\downarrow2.29mm$). However, SPM will degrade the segmentation performance of cervical vertebrae, with the Dice score decreasing by $7.48\%$. We argue that there are 220 cervical, 884 thoracic, 621 lumbar vertebrae in the VerSe 2019 dataset \cite{sekuboyina2021verse}, in which cervical vertebrae make up a small proportion of all vertebrae. Therefore, there might be a risk of biased learning, where explicit shape priors focus on the leaning of thoracic and lumbar vertebrae. Even if a segmentation degradation in the cervical region, average evaluation metrics for the whole spline is still improved, with a $1.23\%$ Dice score increase and a $1.05mm$ $HD_{95}$ decrease. Besides, Figure \ref{ablation visualizations} shows more consistent predictions in each vertebra, which is a strong proof that shape priors can enhance models' representation abilities via the introduction of shape constraints.

\begin{table}[!t]
  \begin{center}
  \caption{Quantitative comparisons between SPM and other attention modules on BraTS 2020, VerSe 2019 and ACDC.}
  \label{attention comparison}
  \resizebox{\columnwidth}{!}{
  \begin{tabular}{ccccccccc}  
  \hline  
  \multirow{2}*{Method} & \multicolumn{2}{c}{BraTS 2020} & \multicolumn{2}{c}{VerSe 2019}  & \multicolumn{2}{c}{ACDC} & \multirow{2}*{Params(M)}  & \multirow{2}*{FLOPs(T)}\\  
  \cmidrule(r){2-3}  \cmidrule(r){4-5}  \cmidrule(r){6-7}
  & Dice $\uparrow$ & $HD_{95}$ $\downarrow$ & Dice $\uparrow$ & $HD_{95}$ $\downarrow$ & Dice $\uparrow$ & $HD_{95}$ $\downarrow$  \\
  \hline
  Baseline + attention gate \cite{oktay2018attention} & 88.77  & 4.54 & 85.47 & 4.19 & 91.67 & 1.23 & 44.21 & 456.98 \\
  Baseline + residual attention gate  & 88.84 & 4.02 & 86.40 & 4.20 & 91.91 & 1.24 & 44.21 & 456.98 \\
  Baseline + SE \cite{hu2018squeeze} & 88.87 & 4.09 & 86.89 & \textbf{1.78} & 91.33 & 1.32 & 20.24 & 354.34 \\

Baseline + CBAM \cite{woo2018cbam} & 88.65 & 3.83 & 85.00 & 4.36 & 91.91 & 1.27 & 22.21 & 386.52 \\
Baseline + SPM & \textbf{89.34} & \textbf{3.63} & \textbf{87.65} & \textbf{1.78} & \textbf{92.34} & \textbf{1.21} & 43.53 & 438.23 \\

  \hline  
  \end{tabular}}
  \end{center}
\end{table}

\subsubsection{Structural Ablations of SPM}
SPM is composed of SUB and CUB, which play a different role in the process of enhancing skipped features and refining explicit shape priors. Thus, we further research on the effectiveness of these two components. According to Table \ref{ablation}, the introduction of the individual CUB brings a significant improvement on the baseline model, with a $0.64\%\uparrow$ Dice score increase and a $1.03mm\downarrow$ $HD_{95}$ decrease on the BraTS 2020, a $1.09\%\uparrow$ average Dice score increase on ACDC. And this is a strong proof that CUB can boost the shape representation for local GT regions. When SUB is applied to the structure of SPM, there is a further performance increase on these two datasets. However, for the VerSe 2019 dataset, removing SUB from SPM will sharply degrade the segmentation performance, even lower than that of the baseline as shown in Table \ref{ablation}. We give an explanation that global contexts from SUB are essential for the identification of vertebrae due to the existence of long-range dependency contained in the longitudinal axis of the spines. Besides, after introducing CUB to the baseline model, the segmentation performance for cervical vertebrae declines significantly while thoracic and lumbar vertebrae are finely segmented, which might result from the biased ratio between three kinds of vertebrae \cite{sekuboyina2021verse}.

\subsubsection{Comparison with other attention mechanisms}
\begin{figure}[!t]
\centerline{\includegraphics[width=\columnwidth]{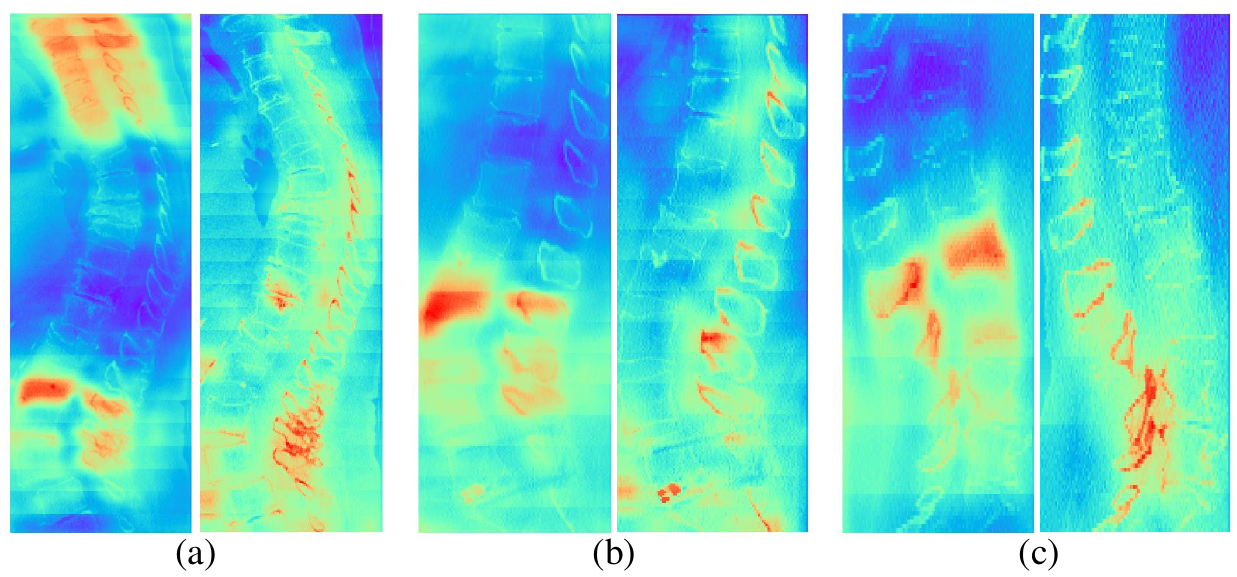}}
\caption{Attention map comparisons between attention gate and SPM. For (a)-(c) cases, the left and right columns represent attention maps from the attention gate and SPM respectively.}
\label{shape priors comparison with attention gate}
\end{figure}

Here SPM is intrinsically a type of attention module to enhance skipped features with luxurious shape information. Thus, we conduct quantitative experiments on the performance comparison with other classic attention modules, popularly employed in the field of medical image segmentation. As shown in Figure \ref{attention comparison}, our proposed attention module can achieve better improvements compared with other attention modules on the three public datasets. Although the parameters and FLOPs of SPM are heavier than those of SE \cite{hu2018squeeze} and CBAM \cite{woo2018cbam} block, there exist significant improvements on the average Dice score due to the fact that our module introduces additional shape information generated from learnable shape priors. Specifically, SPM outperforms SE block on VerSe 2019 and ACDC with a $0.76\%$ and $1.01\%$ Dice score increase, shows superior to CBAM block on BraTS2020 and VerSe 2019 with a $0.69\%$ and $2.65\%$ Dice score increase. Apart from that, the attention gate from attention UNet \cite{oktay2018attention} bears the same form for outputs, with a refined skipped feature and an attention map for the target region. Thus, we give a qualitative visualization for attention maps between the attention gate and SPM. As illustrated by Figure \ref{shape priors comparison with attention gate}, shape priors generated from SPM show stronger localization abilities than attention maps from the attention gate. More detailedly, the latter attention can cover regions of the whole CT spine, which proves that SPM can model global dependency due to the existence of SUB.

\subsubsection{Discussions}
Quantitative and qualitative experimental results on BraTS 2020, VerSe 2019 and ACDC validate the effectiveness of SPM. Besides, as shown in Figure \ref{shape priors}, explicit shape priors show the potential for providing sufficient shape information to guide the segmentation task. Currently, there exists a phenomenon called repetitive activation, which means that different channels of explicit shape priors tend to activate the same region. Here the affinity map $S_{map}$ in the self-update block is employed to describe inter-relations between different channels of shape priors. And we expect to attain shape priors, in which each channel shows discriminative shape priors. Due to the fact that no constraint conditions are imposed to guide the learning of affinity map $S_{map}$, there still exists dependency relations between channels of global shape priors, which results in the repetitive activation in the last stage of shape priors. Furthermore, on the ground that global priors generated from SUB contain global contexts and textures for the whole region, the last-stage shape priors of some cases show less accurate shape information than those in the second stage. As a result, the design for SUB will be a potential direction that needs to be further researched.

\section{Conclusion}
In this paper, we detailedly discuss three types of segmentation models with shape priors, which consist of atlas-based models, statistical-based models and UNet-based models. To enhance the interpretability of shape priors on UNet-based models, we proposed a shape prior module (SPM), which could explicitly introduce shape priors to promote the segmentation performance on different datasets. And our model achieves state-of-the-art performance on the datasets of BraTS 2020, VerSe 2019 and ACDC. Furthermore, according to quantitative and qualitative experimental results, SPM shows a good generalization ability on different backbones, which can serve as a plug-and-play structure.

\bibliographystyle{IEEEtran}
\bibliography{paper}

\end{document}